# AN AGENT-BASED SIMULATION OF IN-STORE CUSTOMER EXPERIENCES

*Dr. Peer-Olaf Siebers*
*Prof. Uwe Aickelin*
University of Nottingham, School of Computer Science (ASAP)
Nottingham, NG8 1BB, UK
pos@cs.nott.ac.uk, uxa@cs.nott.ac.uk

*Ms. Helen Celia*
*Prof. Chris W. Clegg*
University of Leeds, Centre for Organisational Strategy, Learning & Change (LUBS)
Leeds, LS2 9JT, UK
h.celia@leeds.ac.uk, c.w.clegg@leeds.ac.uk

**ABSTRACT:**

*Agent-based modelling and simulation offers a new and exciting way of understanding the world of work. In this paper we describe the development of an agent-based simulation model, designed to help to understand the relationship between human resource management practices and retail productivity. We report on the current development of our simulation model which includes new features concerning the evolution of customers over time. To test some of these features we have conducted a series of experiments dealing with customer pool sizes, standard and noise reduction modes, and the spread of the word of mouth. Our multi-disciplinary research team draws upon expertise from work psychologists and computer scientists. Despite the fact we are working within a relatively novel and complex domain, it is clear that intelligent agents offer potential for fostering sustainable organisational capabilities in the future.*

## 1. INTRODUCTION

The retail sector has been identified as one of the biggest contributors to the productivity gap, whereby the productivity of the UK lags behind that of France, Germany and the USA (Department of Trade and Industry, 2003; Reynolds et al., 2005). There is no doubt that management practices are linked to an organisation's productivity and performance (Wall and Wood, 2005). Operations Research (OR) is applied to problems concerning the conduct and co-ordination of the operations within an organisation (Hillier and Lieberman, 2005). An OR study usually involves the development of a scientific model that attempts to abstract the essence of the real problem. Most standard OR methods can only be used once practices have been implemented and most of them only report on snapshots in time and are therefore predictive tools and less useful if one is interested in understanding the behaviour of a system rather than only predicting its performance.

Simulation can be used to analyse the operation of dynamic and stochastic systems showing their development over time. There are many different types of simulation, each of which has its specific field of application. Agent-Based Simulation (ABS) is particularly useful when complex interactions between system entities exist such as autonomous decision making or proactive behaviour. ABS shows how micro-level processes affect macro level outcomes; macro level behaviour is not explicitly modelled, it emerges from the micro-decisions made by the individual entities (Pourdehnad et al., 2002).

In our research project we investigate how ABS can help with assessing and optimising the impact of Human Resource (HR) management practices on customer satisfaction and the performance of service-oriented retail organisations. For this purpose we have developed a Management Practice Simulation (ManPraSim) model. So far we have only studied the impact of HR management practices (e.g. training and empowerment) on a customer base that is not influenced by any external or internal stimuli - and hence does not evolve (Siebers et al. 2007a; Siebers et al. 2007b). In order to be able to investigate the impact of management practices on customer satisfaction in a more realistic way we need to consider the factors that stimulate customers to respond to these practices. Therefore our focus is currently on building capabilities to model customer evolution as a consequence of the implementation of management practices. Changes in the behaviour of customers over time can be driven by external factors such as a friend's recommendation or internal factors such as memory of one's own previous shopping experiences. Changing customer requirements may in turn alter what makes a successful management practice (as these are context specific, and customers are a key component of any retail context). In order to enable such studies we had to enhance our existing ManPraSim model v1 and introduce some new features.

In this paper we discuss the key features we have implemented in order to allow the investigation of these kinds of behavioural dynamics. What we are learning here about modelling human behaviour has implications for modelling any complex system that involves many human interactions and where the actors work with some degree of autonomy.

## 2. BACKGROUND

Researchers from various disciplines have examined the link that exists between management practices and productivity. Management researchers have published evidence supporting a relationship (Delaney and Huselid, 1996), although others hold a more cynical view (Wall and Wood, 2005). As with any complex human system, it is hard to conclusively delineate the effects of management practices from other socially embedded factors.

There has been a lot of modelling and simulation of operational management practices, but HR management practices have often been neglected although research suggests that they are the largest, least-well understood contributor to the productivity gap (Birdi et al., 2006). One reason for this relates to the key component of HR management practices, an organisation's people, who are often unpredictable in their individual behaviour.

Previous research into retail productivity has typically focused on consumer behaviour and efficiency evaluation (e.g. Patel and Schlijper, 2004; Nicholson et al., 2002), and we seek to build on this work and address the neglected area of HR retail management practices (Keh et al., 2006). Notwithstanding the previous point, in that we are covering new ground, it is important that we build on existing work on the impact of customer behaviour in order to create a highly sophisticated model.

When investigating the behaviour of complex systems the choice of an appropriate modelling technique is very important (Robinson, 2004). In OR a wide variety of modelling approaches are in use. These approaches can be classified into three main categories: analytical, heuristic, and simulation. Often a combination of these is used within a single model (e.g. Greasley, 2005; Schwaiger and Stahmer, 2003). After a thorough investigation of the relevant literature we have identified simulation as being the most appropriate approach for our purposes.

Simulation introduces the possibility of a new way of thinking about social and economic processes, based on ideas about the emergence of complex behaviour from relatively simple activities (Simon, 1996). It allows the testing and evaluation of a theory, and investigation of its implications. There are many different approaches to OR simulation, amongst them Discrete-Event Simulation (DES), System Dynamics (SD), and ABS (sometimes referred to as individual-based simulation) (Borshchev and Filippov, 2004). The choice of the most suitable approach depends on the issues investigated, the input data available, the required level of analysis, and the type of answers that are sought (Robinson, 2004).

Although computer simulation has been used widely since the 1960s, ABS only became popular in the early 1990s (Epstein and Axtell, 1996). It is described by Jeffrey (2007) as a mindset as much as a technology: 'It is the perfect way to view things and understand them by the behaviour of their smallest components'. In ABS a complex system is represented by a collection of agents that are programmed to follow simple behavioural rules. Agents can interact with each other and with their environment to produce complex collective behavioural patterns. The main characteristics of agents are their autonomy, their ability to take flexible action in reaction to their environment, and their pro-activeness depending on motivations generated from their internal states. They are designed to mimic the attributes and behaviours of their real-world counterparts. ABS is still a relatively new simulation technology and its principle application has been in academic research. With the appearance of more sophisticated modelling tools in the broader market, things are starting to change (Luck et al., 2005). Also, an ever increasing number of computer games use the ABS approach.

Due to the characteristics of the agents, this modelling approach appears to be more suitable than DES for modelling human-centred systems as it supports autonomy and pro-activeness. ABS seems to promote a natural form of modelling these systems. There is a structural correspondence between the real system and the model representation, which makes them more intuitive and easier to understand than for example a system of differential equations as used in SD. Hood (1998) emphasises that one of the key strengths of ABS is that the system as a whole is not constrained to exhibit any particular behaviour as the system properties emerge from its constituent agent interactions. Consequently, assumptions of linearity, equilibrium and so on, are not needed.

On the other hand, there is consensus in the literature that it is difficult to evaluate agent-based models, because the behaviour of the system emerges from the interactions between the individual entities. This concern can be addressed by progressively increasing the level of complexity within the model during the design stage or by controlling the noise (taking out stochasticity where ever possible) during the validation process. Furthermore, problems often occur through a lack of adequate empirical data; it has been questioned whether or not a model can be considered to scientifically represent a system when it is not built with 100% objective, and measurable data. However, many of the variables built into a system cannot be objectively quantified but expertly-validated estimates offer a unique solution to this problem. Finally, there is the danger that people new to ABS may expect too much from the models, particularly with regard to predictive ability. To mitigate this problem it is important to be clear with individuals about what this modelling technique can really offer, to guide realistic expectations.

In conclusion we can say that ABS offers a fresh opportunity to realistically and validly model organisational characters and their interactions, which in turn facilitate a meaningful investigation of management practices and their impact on system outcomes.

## 3. MODEL DESIGN AND DATA COLLECTION

Before building a simulation model one needs to understand the particular problem domain (Chick, 2006). In order to gain this understanding we have conducted some case studies. What we have learned during those case studies is reflected in the conceptual models presented in this chapter. Furthermore we explain how we intend to use the data we have gathered during our case studies.

### 3.1 KNOWLEDGE GATHERING

Case studies were undertaken in four departments across two branches of a leading UK retailer. The case study work involved extensive data collection techniques, spanning: participant observation, semi-structured interviews with team members, management and personnel, completion of survey questionnaires and the analysis of company data and reports (for further information, see Celia, 2007). Research findings were consolidated and fed back (via report and presentation) to employees with extensive experience and knowledge of the four departments in order to validate our understanding and conclusions. This approach has enabled us to acquire a valid and reliable understanding of how the real system operates, revealing insights into the working of the system as well as the behaviour of and interactions between the different actors within it. As the operational case study data are confidential they have only been compiled for an internal report and not published.

In order to make sure that our results regarding the application of management practices are applicable for a wide variety of departments we have chosen two different types of case study departments which are substantially different not only in their way of operating but also their customer type split and staff setup. We collected our data in the Audio & Television (A&TV) and the WomensWear (WW) departments of the two case study branches.

The key differences between these two department types can be summarised as follows. The average customer service time in A&TV is significantly longer, and the average purchase is significantly more expensive than in WW. The likelihood of a customer seeking help in A&TV is also much higher than in WW. Out of customers who have received advice, those in WW have a higher likelihood of making a purchase (indeed customers' questions tend to be very specific to a desired purchase) than in A&TV. Considering customer types, A&TV tends to attract more solution demanders and service seekers, whereas WW customers tend to be shopping enthusiasts. Finally, it is important to note that the conversion rate (the likelihood of customers making a purchase) is higher in WW than in A&TV.

## 3.2 CONCEPTUAL MODELLING

We have used the knowledge gained from the case studies to develop our conceptual models of the system to be investigated, the actors within the system, and their behavioural changes due to certain stimuli.

### 3.2.1 Main Concepts for the Simulation Model

Our initial ideas for the simulation model and its components are shown in Figure 1. Regarding system input we use different types of agents (customers, sales staff and managers), each with a different set of relevant attributes and we have some global parameters which influence any aspect of the system. The core of our ManPraSim model consists of an ABS model with a user interface to allow some form of user interaction (change of parameters) before and during runtime. Regarding system outputs, we aim to find some emergent behaviour on a macro level. Visual representation of the simulated system and its actors allows us to monitor and better understand the interactions of entities within the system. Coupled with the standard DES performance measures, we can then identify bottlenecks to assist with optimisation of the modelled system.

### 3.2.2 Concepts for the Actors

We have used state charts for the conceptual design of our agents. State charts show the different states an entity can be in and define the events that cause a transition from one state to another. This is exactly the information we need in order to represent our agents at a later stage within the simulation environment. We have found this form of graphical representation a useful part of the agent design process because it is easier for an expert in the real system (who is not an expert in ABS) to quickly take on board the model conceptualisation and provide useful validation of the model structure and content.

Designing and building a model is to some extent subjective, and the modeller has to selectively simplify and abstract from the real scenario to create a useful model (Shannon, 1975). A model is always a restricted copy of the real world, and an effective model consists of only the most important components of the real system. In our case, the important system components take the form of the behaviours of an actor and the triggers that initiate a change from one behaviour to another. We have developed state charts for all the relevant actors in our simulation model. Figure 2 shows as an example the conceptual model of our customer agents.

### 3.2.3 Concepts for a Novel Performance Measure

We introduce a service level index as a novel performance measure using satisfaction weights. Historically customer satisfaction has been defined and measured in terms of customer satisfaction with a purchased product (Yi, 1990). The development of more sophisticated measures has moved on to incorporate customers' evaluations of the overall relationship with the retail organisation, and a key part of this is the service interaction. Indeed, empirical evidence suggests that quality is more important for customer satisfaction than price or value-for-money (Fornell et al., 1996), and extensive anecdotal evidence indicates that customer-staff service interactions are an important determinant of quality as perceived by the customer.

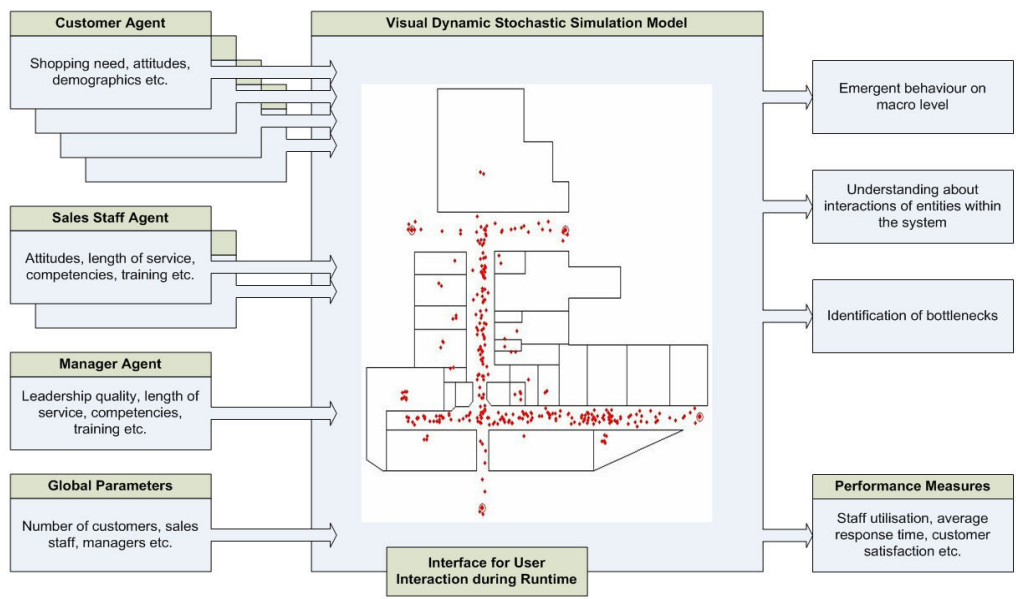

Figure 1: Initial ideas for the simulation model and its components

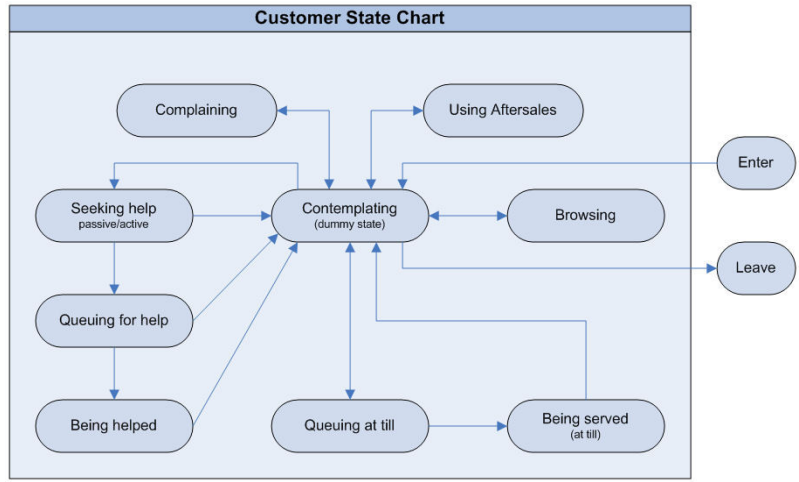

Figure 2: Conceptual model of our customer agents (transition rules have been omitted for clarity)

The index allows customer service satisfaction to be recorded throughout the simulated lifetime. The idea is that certain situations might have a bigger impact on customer satisfaction than others, and therefore weights can be assigned to events to account for this. Applied in conjunction with an ABS approach, we expect to observe interactions with individual customer differences; variations which have been empirically linked to differences in customer satisfaction (e.g. Simon and Usunier, 2007). This helps the analyst to find out to what extent customers underwent a positive or negative shopping experience and it also allows the analyst to put emphasis on different operational aspects and try out the impact of different strategies.

### 3.2.4 Concepts for Modelling Customer Evolution

There are two different ways in which we consider customer evolution: external stimulation attributable to Word Of Mouth (WOM) and internal stimulation triggered by memory of one's own previous shopping experiences (this is still work in progress). Sharing information with other individuals (referred to as WOM), significantly affects the performance of retail businesses (Marsden et al., 2005). An important source of WOM results from customer experiences of retail outlets, and a customer's judgement about whether or not the experience left them feeling satisfied (or dissatisfied). We incorporate WOM in our simulation model by using the number of satisfied customers at the end of the day to calculate the number of additional customers visiting on the next day. The calculation takes into account that only a fraction of people act upon received WOM. Our concept for representing internal stimulation comprises the exertion of influence on picking certain customer types more often than others. An enthusiastic shopper with a high satisfaction score is much more likely to go shopping more frequently than a disinterested shopper. Therefore, we are introducing some constraints (e.g. out of all customers picked 50% have to be enthusiastic shoppers, 30% normal shoppers, and 20% disinterested shoppers, or, customers with a higher satisfaction score are more likely to be picked to revisit the department).

### 3.3 EMPIRICAL DATA

Often agents are based on analytical models or heuristics and, in the absence of adequate empirical data, theoretical models are employed. However, we use frequency distributions for modelling state change delays and probability distributions for modelling decision making processes because statistical distributions are the best way in which we can represent the numerical data we have gathered during our case study work. In this way a population is created with individual differences between agents, mirroring the variability of attitudes and behaviours of their real human counterparts.

The frequency distributions are modelled as triangular distributions defining the time that an event lasts, using the minimum, mode, and maximum duration and these figures are based on our own observations and expert estimates in the absence of objective numerical data.

The probability distributions are partly based on company data (e.g. the rate at which each shopping visit results in a purchase) and partly on informed estimates (e.g. the patience of customers before they leave a queue). Table 1 and 2 show some of the distributions we have defined for our simulation models. We also gathered some company data about work team numbers and work team composition, varying opening hours and peak times, along with other operational details.

## 4. IMPLEMENTATION

### 4.1 IMPLEMENTATION OF THE MAIN CONCEPTS

Our ManPraSim model has been implemented in AnyLogic™ (version 5.5) which is a Java™ based multi-paradigm simulation software (XJ Technologies, 2007). During the implementation we have applied the knowledge, experience and data accumulated through our case study work. Within the simulation model we can represent the following actors: customers, service staff (with different levels of expertise) and managers. Figure 3 shows a screenshot of the customer and staff agent logic in AnyLogic™ as it has been implemented in the latest version of our simulation model. Boxes represent states, arrows transitions, arrows with a dot on top entry points, circles with a B inside branches, and numbers satisfaction weights.

At the beginning of each simulation run a customer pool is created which represents a population of potential customers that might visit the simulated department on an unspecified number of occasions. Once the simulation has started customers are chosen at a specified rate (customer arrival rate) and released into the simulated department. Currently two different customer types are implemented: customers who want to buy something and customers who require a refund. If a refund is granted, a customer decides whether his or her goal changes to leaving the department straight away, or to making a new purchase. The customer agent template consists of four main blocks which all use a very similar logic. In each block, in the first instance, customers try to obtain service directly and if they cannot obtain it (i.e. no suitable staff member is available) they have to queue. They then either be served as soon as the suitable staff member becomes available, or leave the queue if they do not want to wait any longer (an autonomous decision). A complex queuing system has been implemented to support different queuing rules. Once customers have finished their shopping (either successfully or not) they leave the simulated department and are added back to the customer pool where they rest until they are picked the next time.

While the customer is in the department a satisfaction score is calculated by summing the satisfaction weights attached to the transitions that take place during the customer's visit. For example, a customer starts browsing and then requires some help. If he or she gets help immediately his or her satisfaction score goes up (+2) and after he or she received the help the score goes up again (+2). He or she then moves to the till. If he or she has to wait for help and leaves the queue because he or she is fed up waiting his or her score goes down (-2). Upon leaving the department he or she will end up with an overall satisfaction score of +2.

In comparison to the customer agent state chart, the staff agent state chart is relatively simple. Whenever a customer requests a service and the staff member is available and has the right level of expertise for the task requested, the staff member commences this activity until the customer releases the staff member. Whereas the customer is the active component of the simulation model, the staff member is currently passive, simply reacting to requests from the customer.

### 4.2 KEY FEATURES

There are some additional key features that the simulation model possesses which we describe in the following two sections. First we provide an overview of some important features which have already been implemented in the previous version of our simulation model (ManPraSim model v2) but are relevant to modelling the evolution of customers. This sets the scene for the subsequent description of the new features we have implemented in the latest version of our simulation model (ManPraSim model v3).

| event | probability it occurs |
|---|---|
| someone makes a purchase after browsing | 0.37 |
| someone requires help | 0.38 |
| someone makes a purchase after getting help | 0.56 |

Table 1: Sample frequency distribution values

| situation | min | mode | max |
|---|---|---|---|
| leave browse state after … | 1 | 7 | 15 |
| leave help state after … | 3 | 15 | 30 |
| leave pay queue (no patience) after … | 5 | 12 | 20 |

Table 2: Sample probability values

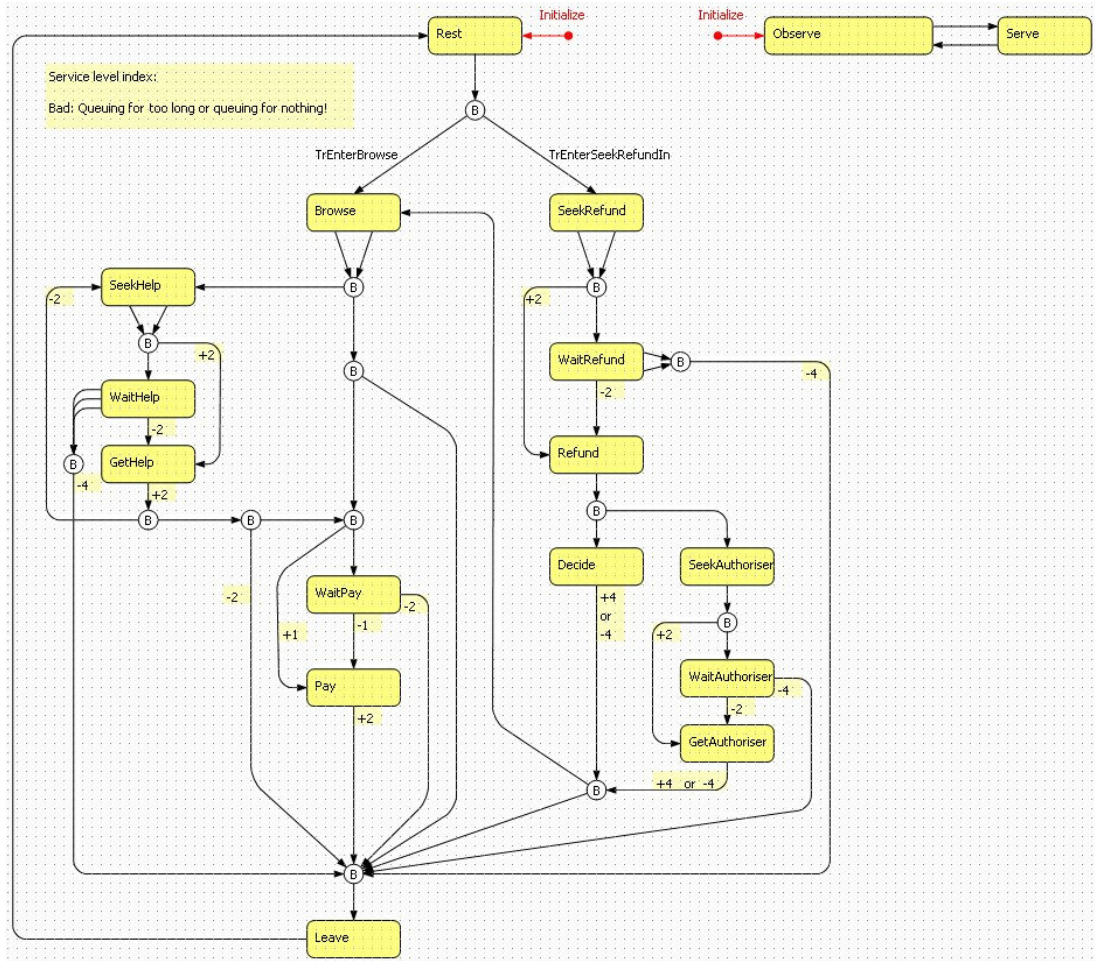

Figure 3: Customer (left) and staff (right) agent logic implementation in AnyLogic™

### 4.2.1 Key Features of the ManPraSim Model v2 (previous version)

In the ManPraSim model v2 we introduced realistic footfall, customer types, a finite population of customers, and a quick exit at shop closing time. In this paper we only provide an overview of these features because they are described in more detail elsewhere (see Siebers et al., 2007c).

**Realistic footfall:** There are certain peak times where the pressure on staff members is higher. In v2 of our simulation model we have implemented these hourly fluctuations through the addition of realistic footfall (based on automatically recorded sales transaction data) reflecting different patterns of customer footfall during the day and across different days of the week. In addition we can model the varying opening hours on different days.

**Customer types:** In real life, customers display certain non-transient shopping preferences and behaviours which can be broadly categorised into a finite number of customer types. To create a realistic and diverse set of customers for our simulation model we have introduced five customer types (shopping enthusiasts, solution demanders, service seekers, disinterested shoppers and internet shoppers). The definition of each customer type is based on values which define the likelihood of performing a certain action (see Table 3).

In the simulation model we have two algorithms which have been developed to imitate the influence of these attributes on customer behaviour. They have been implemented as methods that are invoked when defining the state change delays modelled by triangular frequency distributions and when supporting decision making modelled by probability distributions. Basically the methods define new threshold values for the distributions based on the likelihood values mentioned above. Program 1 shows as an example the pseudo code for the probability distribution threshold correction algorithm. If the customer is a shopping enthusiast and is about to make the decision whether to make a purchase or to leave the department directly (see Figure 3, customer state chart, second branch after leaving the browse state) a corrected threshold value (probability) for this decision is calculated. For this calculation the original threshold of 0.37 (see Table 2) is taken into account and, as for shopping enthusiasts the likelihood to buy is high (see Table 3), the corrected threshold value is calculated as follows: 0.37+0.37/2 = 0.56. Consequently the likelihood that the shopping enthusiast proceeds to the checkout rather than leaving the department without making any purchase has risen by 18.5%.

| Customer type | Likelihood to | | | |
|---|---|---|---|---|
| | buy | wait | ask for help | ask for refund |
| Shopping enthusiast | high | moderate | moderate | low |
| Solution demander | high | low | low | low |
| Service seeker | moderate | high | high | low |
| Disinterested shopper | low | low | low | high |
| Internet shopper | low | high | high | low |

Table 3: Definitions for each type of customer

**Finite customer population:** A key aspect to consider is that the most interesting system outcomes evolve over time and many of the goals of the retail company (e.g. service standards) are planned strategically over the long-term. We have therefore introduced a finite population of customers (customer pool) where each customer agent is given a certain characteristic based on the customer types mentioned above. The customer type split in the customer pool can be defined via an initialisation file before the execution of the simulation. The shopping experience of each visit (satisfaction index) is stored in the long term memory of the agent after he or she has left the department. In this way the service a customer experiences can be evaluated over the complete simulated time span.

**Quick exit at closing time:** We have added transitions that emulate the behaviour of customers when the store is closing. These transitions are immediate exits of each customer's current state (i.e. the equivalent to a customer running out of shopping time and leaving the store). Not all customer states have these additional transitions as it is for example very unlikely that customers leave the store immediately when they are already queuing to pay. Now the simulated department empties within a ten to fifteen minute period, which conforms to what we have observed in the real system.

### 4.2.2 Key Features of the ManPraSim Model v3 (current version)

In the ManPraSim model v3 we have introduced a staff pool with an additional staff type, a new operation mode to support sensitivity analyses and a first implementation of modelling WOM. Furthermore, we have created some new performance measures that allow us to measure and record the shopping experience of individual visits to the department as well as the daily performance of the department.

**Staff pool and additional staff types:** Retail trends reflect that shops are now open for longer hours over more days of the week, and our case study organisation is no exception. To accurately incorporate this source of system variability, we have introduced a staff pool to allow different staffing on different days of the week. The

```
for (each threshold to be corrected) do
{
  if (OT < 0.5) limit = OT / 2 else (limit = 1 - OT) / 2
  if (likelihood = 0) CT = OT - limit
  if (likelihood = 1) CT = OT
  if (likelihood = 2) CT = OT + limit
}

where: -    OT = original threshold
            CT = corrected threshold
            likelihood: 0 = low, 1 = moderate, 2 = high
```

Program 1: Pseudo code for the probability distribution threshold correction algorithm

simulation uses Full Timers (FT) to cover all staff shifts required during weekdays. Additional staff who are required to cover busy weekend shifts are modelled by Part Timers (PT). The maximum number of staff required of each type is calculated during the simulation initialisation. A staff pool is then created to include weekday FT staff of different types, and the required generic PT to fill the gaps left in the staff shifts which need to be covered. At the beginning of each day the simulation checks how many staff are required and picks the required amount of staff out of the pool at random. The selection process is ordered as follows: FT first and then PT, if required. PT staff have been defined as a generic staff type and can take over any required role. We have tried to model a staff rota with more complex constraints, for example FT staff working five days followed by two days off, however this has as yet proved unsuccessful. Therefore the currently modelled shifts do not incorporate days off work for FT staff.

**Noise reduction mode:** A noise reduction mode has been implemented which allows us to conduct a sensitivity analysis with constant customer arrival rates, constant staffing throughout the week and constant opening hours. We have used the average values of real world case study data to define the constant values. This has resulted in different values for the different case study department types. As this is a simulation model we cannot take out all the stochasticity, but this way at least we can reduce the system noise to clearly see the impact of the parameter under investigation. When we progress to reintroduce the system noise, the knowledge we have accumulated when experimenting in noise reduction mode helps us to better understand patterns in systems outcomes, and be better able to attribute causation to the introduction of a particular variable (bearing in mind that in the end we are particularly interested in patterns between variables when they are interacting with one another - not in isolation).

**Word of mouth:** We have developed different strategies to implement WOM which we will test one after the other. Our current algorithm works as follows. We count the number of customers per day who are satisfied and those who are dissatisfied with the service provided during their shopping visit. Satisfied customers are likely to recommend shopping in the department to others whereas dissatisfied customers are likely to advise others not to visit. The adoption rate (i.e. the success rate of convincing others to commit to either action) depends on the number of people that act upon the received WOM (adoption fraction) and how many contacts a customer has (contact rate). The Equation for this calculation is shown in Equation 1.

In our current algorithm we simply pick the additional customers out of our customer pool at random. In a later implementation we want to expand or decrease our customer population rather than using the existing customers to model WOM influence. This seems to be closer to reality as the WOM in most cases carries positive messages and therefore attracts new visitors rather than motivating existing customers to come more often. Negative WOM motivates potential customers not to visit the department in the first place.

**New performance measures:** With the introduction of a finite population (represented by our customer pool) we have had to rethink the way in which we collect statistics about the satisfaction of customers. Previously, the life span of a customer has been a single visit to the department. At the end of his or her visit, the individual's satisfaction score (direction and value) has been recorded. Now the life span of a customer lasts the full runtime of the simulation and he or she can be picked several times to visit the department during that period. Our previous performance measures now collect different information: satisfaction scores considering customers' satisfaction history. These measures do not reflect individuals' satisfaction with the current service experience but instead the satisfaction with the overall service experience during the lifetime of the agent. Furthermore, they are biased to some extent in that an indifferent rating quickly shifts into satisfaction or dissatisfaction (arguably this is realistic because most people like to make a judgement one way or the other). Whilst this is still a valuable piece of information we would also like to know how current service is perceived by each customer. For this reason we have introduced a set of new performance measures to record the experience of

---

$$n_{\text{additional customers}(d)} = (n_{\text{satisfied}(d-1)} - n_{\text{dissatisfied}(d-1)}) * \text{adoption fraction} * \text{contact rate}$$

where: -  d = current day
       d-1 = previous day
       n = number of …

Equation 1: Calculation of the adoption rate

each customer's individual visit. Basically these are the same measures as before but on a day-to-day basis they are not anchored by the customer's previous experiences. We examine the sum of these 'per visit' scores across the lifetime of all customers. Another new measure tracks the satisfaction growth for customers' current and overall service experience. With the incorporation of varying customer arrival rates, opening hours and staffing we have brought in a set of performance measures that capture the impact of these variations on a daily basis. These measures are particularly useful for optimising departmental performance throughout the week. We also record the satisfaction growth per each individual customer visit. At the end of the simulation run the simulation model produces a frequency distribution which informs us about how satisfied or dissatisfied individual customers have been with the service provided. Furthermore, all forms of customer queue (till, normal help, expert help, and refund decision) are now monitored through new performance measures that record how many people have been queuing in a specific queue, and how many of these lost their patience and left the queue prematurely. This measure helps us to understand individual customers' needs because it tells us what individual customers think about the service provided. Finally, we have added some methods for writing all parameters and performance measures into files to support documentation and analysis of the experiments. Some problems have arisen with our utilisation measures since the introduction of PTs. These workers can take on any role and therefore cannot easily be attributed to a specific utilisation statistic. This problem still needs to be resolved but it only affects the utilisation measures when the simulation runs in normal mode; most experiments use the noise reduction mode.

### 4.3 Model Validation

Validation ensures that the model meets its intended requirements in terms of the methods employed and the results obtained. In order to test the operation of the ManPraSim model v3 and ascertain face validity we have completed several experiments. It has turned out that conducting the experiments with the data we collected during our case study did not satisfactorily match the performance data of the real system. We identified the staff setup used in the simulation models as being the main cause of the problem. The data we had used here had been derived from real staff rotas. On paper these real rotas suggested that all workers are engaged in exactly the same work throughout the day but we know from working with, and observing workers in, the case study organisation that in reality each role includes a variety of activities. Staff members in the real organisation allocate their time between competing tasks such as customer service, stock replenishment, and taking money. So far our simulation models incorporate only one type of work per staff member. For example, the A&TV staff rota indicates that only one dedicated cashier works on weekdays. When we have attempted to model this arrangement, customer queues became extremely long, and the majority of customers ended up losing their patience and leaving the department prematurely with a high level of dissatisfaction. In the real system we observed other staff members working flexibly to meet the customer demand, and if the queue of customers grew beyond a certain point then one or two would step in and open up further tills to take customers' money before they became dissatisfied with waiting. Furthermore, we observed that a service staff member, when advising a customer, would often continue to close the sale (filling in guarantee forms and taking the money off the customer) rather than asking the customer to queue at the till for a cashier whilst moving on to the next customer.

This means that currently our abstraction level is too high and we do not model the real system in an appropriate way. We hope to be able to fix this in a later version. For now we do not consider this to be a big problem so long as we are aware of it. We model as an exercise to gain insights into key variables and their causes and effects and to construct reasonable arguments as to why events can or cannot occur based on the model; we model for insights, not precise numbers.

In our experiments we have modulated the staffing levels to allow us to observe the effects of changing key variables but we have tried to maintain the main characteristic differences between the departments (i.e., we still use more staff in the WW department compared to the A&TV department, only the amount has changed).

### 5. EXPERIMENTS

The purpose of the experiments described below is to further test the behaviour of the simulation model as it becomes increasingly sophisticated, rather than to investigate management practices per se. We have defined a set of standard settings (including all probabilities, staffing levels and the customer type split) for each department type which we use as a basis for all our experiments. The standard settings exclude the customer pool size which are determined in the first experiment and added to the set of standard settings for subsequent experiments. For each experiment the run length has been 10 weeks and we have conducted 20 replications to allow rigorous statistics.

## 5.1 COMPARING CUSTOMER POOL SIZES

The first experiment is a sensitivity analysis. We want to find out what impact the customer pool size has on the simulation results and which of our performance measures are affected by it. For this experiment we have used our new noise reduction mode (described in Section 4.2.2). It allows us to focus our attention on the impact of the variable to be investigated. We have run the experiments for both case study department types using the standard settings and a pool size ranging from 2,000 to 10,000 customers in increments of 2,000.

Our first hypothesis is that varying the customer pool size will not significantly change standard customer measures upon leaving the department. Secondly, we hypothesise that the customer satisfaction measures which accumulate historical data (CSM-AHD) will result in more customers falling into the 'satisfied' and 'dissatisfied' categories compared to the customer satisfaction measure which records the experience per visit (CSM-EPV). Our third hypothesis is that CSM-AHD will interact with the customer pool size while CSM-EPV will not.

Our results (see Tables 4) clearly support all hypotheses. For this experiment we have conducted no further statistics as results are clear-cut. Inspection of the descriptives for hypothesis 1 reveals that ratings for the standard performance measures (leaving after making a purchase, leaving before receiving normal or expert help, leaving whilst waiting to pay, leaving before finding anything) are approximately the same across different customer pool sizes. We predicted this pattern because customer decisions are driven by customer types and the proportional mix of each type is kept constant throughout the experiment. Looking at results for hypothesis 2, there are significant differences between the two different types of satisfaction measures. As expected, CSM-AHD results in a significantly greater number of satisfied and dissatisfied customers and a significantly smaller number of neutral customers than CSM-EPV. This phenomenon has been discussed in Section 4.2.2. Inspection of the descriptives for hypothesis 3 shows that CSM-AHD varies for different customer pool sizes while CSM-EPV stays relatively constant. Furthermore we can see that the bigger the customer pool size the more the CSM-AHD values converge to the CSM-EPV values. This can be explained by the fact that with a larger population the likelihood that a specific customer enters the department repeatedly (and therefore accumulates some historical data) is getting smaller. If we would make the customer population large enough we could expect both customer satisfaction measures to show similar results as we could expect the majority of customers only to be picked once during the simulation runtime and therefore to enter the department with a neutral satisfaction score rather than with an accumulated score.

| A&TV | | | | | | | | | | |
|---|---|---|---|---|---|---|---|---|---|---|
| Customer pool size | 2,000 | | 4,000 | | 6,000 | | 8,000 | | 10,000 | |
| Customers … | Mean | SD | Mean | SD | Mean | SD | Mean | SD | Mean | SD |
| … leaving after making a purchase | 29.5% | 0.1% | 29.5% | 0.1% | 29.6% | 0.1% | 29.5% | 0.1% | 29.4% | 0.1% |
| … leaving before receiving normal help | 2.7% | 0.2% | 2.7% | 0.2% | 2.8% | 0.2% | 2.7% | 0.2% | 2.8% | 0.2% |
| … leaving before receiving expert help | 1.1% | 0.1% | 1.1% | 0.1% | 1.1% | 0.0% | 1.1% | 0.0% | 1.1% | 0.0% |
| … leaving whilst waiting to pay | 17.5% | 0.3% | 17.4% | 0.3% | 17.3% | 0.4% | 17.4% | 0.3% | 17.4% | 0.3% |
| … leaving before finding anything | 49.3% | 0.4% | 49.3% | 0.4% | 49.3% | 0.3% | 49.3% | 0.3% | 49.3% | 0.3% |
| … leaving satisfied (accumulated historical data) | 46.5% | 1.9% | 45.3% | 1.2% | 43.8% | 0.8% | 42.9% | 0.6% | 42.0% | 0.9% |
| … leaving neutral (accumulated historical data) | 8.7% | 0.2% | 12.1% | 0.3% | 14.6% | 0.2% | 16.5% | 0.2% | 18.1% | 0.3% |
| … leaving dissatisfied (accumulated historical data) | 44.8% | 2.2% | 42.5% | 1.4% | 41.6% | 1.1% | 40.6% | 1.0% | 39.9% | 1.1% |
| … leaving satisfied (experience per visit) | 36.8% | 0.3% | 36.9% | 0.2% | 36.9% | 0.2% | 36.8% | 0.2% | 36.7% | 0.4% |
| … leaving neutral (experience per visit) | 35.3% | 0.3% | 35.3% | 0.3% | 35.3% | 0.3% | 35.2% | 0.3% | 35.2% | 0.2% |
| … leaving dissatisfied (experience per visit) | 27.9% | 0.6% | 27.8% | 0.5% | 27.8% | 0.6% | 27.9% | 0.5% | 28.0% | 0.6% |
| WW | | | | | | | | | | |
| Customer pool size | 2,000 | | 4,000 | | 6,000 | | 8,000 | | 10,000 | |
| Customers … | Mean | SD | Mean | SD | Mean | SD | Mean | SD | Mean | SD |
| … leaving after making a purchase | 46.5% | 0.1% | 46.5% | 0.1% | 46.5% | 0.1% | 46.6% | 0.1% | 46.5% | 0.1% |
| … leaving before receiving normal help | 0.0% | 0.0% | 0.0% | 0.0% | 0.0% | 0.0% | 0.0% | 0.0% | 0.0% | 0.0% |
| … leaving before receiving expert help | 0.1% | 0.0% | 0.1% | 0.0% | 0.1% | 0.0% | 0.1% | 0.0% | 0.1% | 0.0% |
| … leaving whilst waiting to pay | 10.0% | 0.2% | 10.0% | 0.3% | 10.0% | 0.3% | 10.0% | 0.2% | 10.0% | 0.2% |
| … leaving before finding anything | 43.4% | 0.3% | 43.4% | 0.3% | 43.3% | 0.3% | 43.3% | 0.2% | 43.3% | 0.2% |
| … leaving satisfied (accumulated historical data) | 93.4% | 0.3% | 88.3% | 0.3% | 84.0% | 0.3% | 80.7% | 0.3% | 78.0% | 0.3% |
| … leaving neutral (accumulated historical data) | 3.9% | 0.1% | 7.2% | 0.2% | 10.2% | 0.2% | 12.6% | 0.2% | 14.7% | 0.2% |
| … leaving dissatisfied (accumulated historical data) | 2.7% | 0.2% | 4.5% | 0.3% | 5.8% | 0.2% | 6.7% | 0.3% | 7.3% | 0.3% |
| … leaving satisfied (experience per visit) | 52.4% | 0.2% | 52.5% | 0.1% | 52.5% | 0.1% | 52.5% | 0.1% | 52.5% | 0.1% |
| … leaving neutral (experience per visit) | 40.3% | 0.3% | 40.3% | 0.2% | 40.2% | 0.3% | 40.2% | 0.2% | 40.2% | 0.2% |
| … leaving dissatisfied (experience per visit) | 7.3% | 0.2% | 7.2% | 0.3% | 7.2% | 0.2% | 7.2% | 0.2% | 7.3% | 0.2% |

Table 4: Descriptives for Experiment 1

The results demonstrate that it is therefore important to select and maintain one customer pool size to ensure that all performance measures are providing comparable information for different experiments. Our case study organisation does not collect or hold data on customer pool size so we have instead calculated a suitable value based on the average numbers of customers who visit the department per day (585 for A&TV and 915 for WW) and an estimate of customers' average inter-arrival time (two weeks for A&TV and one week for WW). These values have been estimated considering the standard customer type split for the corresponding department as well as customer demand for the items sold in that department. They do not necessarily apply to other customer type splits. Using these values we have calculated an appropriate customer pool size for each department (8,000 for A&TV and 6,500 for WW). We use these customer pool sizes for all subsequent experiments.

## 5.2 COMPARING NORMAL AND NOISE REDUCTION MODE

In our second experiment we want to investigate the importance of considering hourly differences in customer arrival rates and daily differences in staffing and opening hours (normal mode). These features have been added to investigate some specific hypotheses related to the real performance of the case study departments and specific patterns that occur on a day-to-day basis. Modelling this level of detail might not be helpful when we conduct a sensitivity analysis where we want to be able to attribute causation to the introduction of a particular variable. For this kind of experiments we would like to be able to control some of the system noise to clearly see the impact of the parameter under investigation (noise reduction mode).

Our fourth hypothesis predicts that in general the runtime performance measures will not be significantly affected by the different level of detail between the two modes (at least for those measures that measure more frequently occurring events). Conversely, our fifth hypothesis asserts that the two modes will produce significantly different values when looking at the daily performance measures on a day-to-day basis (rather than using averages).

Our results provide some support for the hypotheses (see Table 5 and Figure 4). For this experiment we have conducted no further statistics as results are clear-cut. Examining the descriptives, in most cases the runtime performance measures in both modes are approximately the same with a small number of exceptions. Contrary to hypothesis 4, in WW approximately 25% more customers leave whilst waiting to pay in the normal mode (which consequently influences both customer satisfaction measures), as opposed to the noise reduction mode. This apparent cashier bottleneck must be exacerbated by any combination of the three factors which are held constant in noise reduction mode. Further analysis is required to isolate the precise cause. We can also see that, as predicted, the smaller the values the more they differ (on an absolute basis) between the two modes as they are accumulated from a smaller number of events and therefore the influence of different random number streams is more apparent.

When looking at the daily measures on a day-to-day basis (Figure 4 shows daily number of customers and transaction for A&TV as an example) there is a clear differentiation between weekdays, Saturdays and Sundays apparent for the normal mode results while the noise reduction mode, as expected, shows no clear patterns. This additional information can be very useful for optimising the system. For example, we have the lowest number of transactions on Sundays (day 1, 8, etc.) although we do not have the lowest number of customers on this day, therefore we must have a problem with the staffing on Sundays, as at average more people leave the shop on Sundays without buying anything, although the probabilities to do so are the same for all days of the week.

| Customer pool size | A&TV | | | | WW | | | |
|---|---|---|---|---|---|---|---|---|
| | Noise Reduction | | Normal Mode | | Noise Reduction | | Normal Mode | |
| Customers … | Mean | SD | Mean | SD | Mean | SD | Mean | SD |
| … leaving after making a purchase | 12,070.05 | 36.59 | 12,480.30 | 48.29 | 29,698.70 | 61.34 | 29,226.70 | 95.87 |
| … leaving before receiving normal help | 1,134.05 | 78.03 | 1,628.95 | 107.09 | 5.10 | 3.34 | 22.90 | 8.01 |
| … leaving before receiving expert help | 451.15 | 15.74 | 369.95 | 17.60 | 81.60 | 7.98 | 81.65 | 11.39 |
| … leaving whilst waiting to pay | 7,038.60 | 118.86 | 6,711.20 | 130.11 | 6,418.00 | 197.89 | 8,668.45 | 117.61 |
| … leaving before finding anything | 20,179.10 | 135.42 | 20,034.90 | 156.72 | 27,738.65 | 181.02 | 27,792.60 | 173.90 |
| … leaving satisfied (accumulated historical data) | 17,603.20 | 339.74 | 18,688.00 | 352.99 | 53,742.15 | 290.38 | 53,854.70 | 296.31 |
| … leaving neutral (accumulated historical data) | 6,792.05 | 90.33 | 6,538.05 | 79.86 | 6,476.30 | 123.37 | 6,614.05 | 126.08 |
| … leaving dissatisfied (accumulated historical data) | 16,477.70 | 452.41 | 15,999.25 | 519.05 | 3,723.60 | 201.43 | 5,323.55 | 279.09 |
| … leaving satisfied (experience per visit) | 15,070.55 | 115.54 | 15,481.90 | 116.61 | 33,547.60 | 104.55 | 33,296.10 | 122.04 |
| … leaving neutral (experience per visit) | 14,441.25 | 114.92 | 14,344.85 | 140.12 | 25,745.30 | 173.89 | 25,733.20 | 163.40 |
| … leaving dissatisfied (experience per visit) | 11,361.15 | 232.18 | 11,398.55 | 286.37 | 4,649.15 | 144.62 | 6,763.00 | 138.79 |

Table 5: Comparing operation modes, descriptive statistics for A&TV and WW

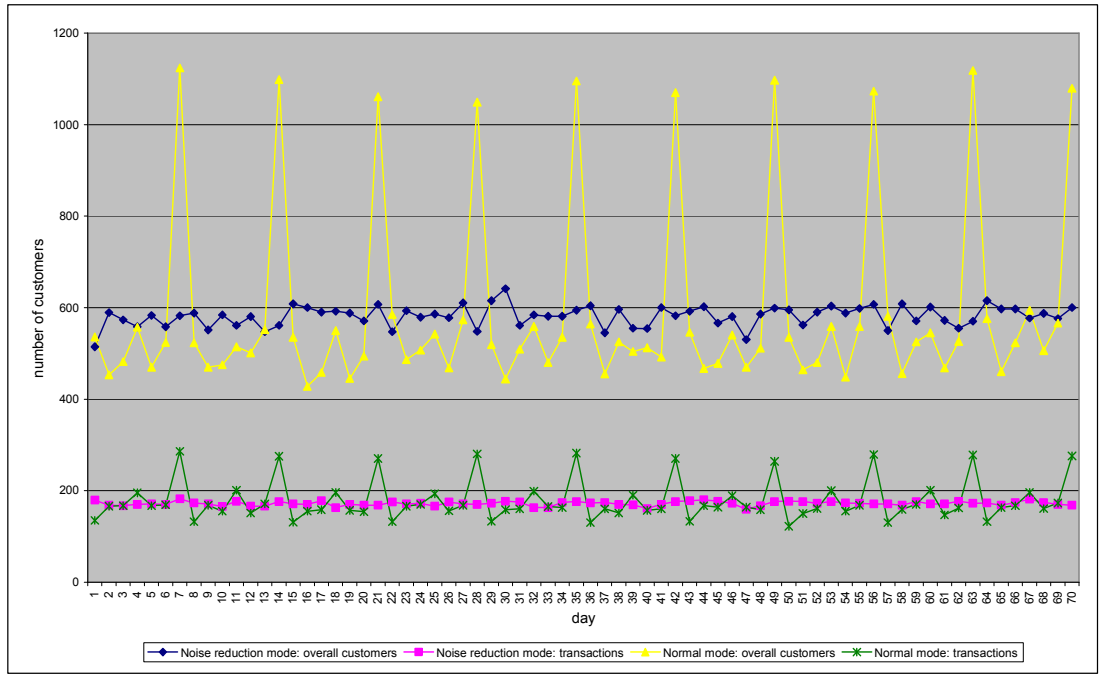

Figure 4: Daily number of customers and transactions for A&TV

Overall the experiment has shown that it is legitimate to use the noise reduction mode within the scope of a sensitivity analysis to test the impact of a specific factor on system behaviour (but we have to keep in mind that measures based on rare events deliver different results in both modes). It is only when we are interested in specific features that are not available in noise reduction mode (e.g. when we want to study differences between particular days of the week or when we need to obtain quantitative data to optimise a real system) that we need to choose the normal mode.

### 5.3 EVALUATING THE IMPLEMENTATION OF WOM

Our final experiment tests our implementation of WOM. We want to find out if the way we have implemented it makes a difference to the performance of the overall system, and whether or not we will observe an interaction with department type. We predict that the higher the WOM adoption fraction, the greater the increase we will see in the number of customer visits. We hypothesise this relationship will be linked to a reduction in satisfaction per customer. This makes sense because the department's staffing resources remain the same and therefore need to be shared between a larger group of customers. We hypothesise that this pattern will vary over time because the impact of WOM will vary on a daily basis.

The results are presented in multiple forms to allow more detailed analysis (see Table 6 and Figure 5). Tabulated values are daily averages across all replications. A series of independent samples T-tests were conducted (see Table 6) to compare department performance measures between the two extreme conditions: an adoption fraction of 0, and an adoption fraction of 1. Levene's equality of variances was violated ($p<.05$) for customer count (A&TV, WW), and customers leaving whilst waiting for normal help (WW only) therefore equal variances have not been assumed for tests of these variables.

T-tests reveal significant differences between all performance measures, with the exception of customers leaving satisfied and before receiving expert help in A&TV. Note however that some effect sizes are only small-to-moderate (eta2-squared =< .06). In A&TV, t-tests reveal a small effect size (eta squared = .04) of resulting in increased customer figures. In WW, a much greater effect size can be observed (eta squared = .66). The number of transactions remains relatively stable over the 10 week period for both departments, which is surprising given the increased number of customer visits. It appears that the restricted availability of staff to serve customers is preventing a commensurate increase in sales. The bottlenecks in A&TV occur with the provision of normal help and cashier availability, whereas the bottlenecks in WW occur to a significant extent with the same variables and also the provision of expert help. Inspection of the mean figures reveals only small absolute differences between customer numbers leaving before receiving normal or expert advice (due to the lesser likelihood these events in WW). All of these bottlenecks result in a significant rise in customers who leave without finding anything to buy in A&TV and WW (eta-squared values of .00 and .30 respectively). The hypothesised relationship with customer satisfaction has not been supported. In A&TV, there is no significant difference (p = .99), and in WW there is a significant but small effect size (eta squared = .03) in the opposite direction to the predicted relationship. We expect that if we looked at other measures of customer satisfaction (i.e. neutrality and dissatisfaction) the pattern we hypothesized would be manifested in these indices.

|  | Adoption fraction | AF = 0 | | AF = 0.5 | | AF = 1 | | T-Test | | |
|---|---|---|---|---|---|---|---|---|---|---|
|  | Customers … | Mean | SD | Mean | SD | Mean | SD | t-value | p-value | Eta² |
| A&TV | Overall customer count [per day] | 584.17 | 6.04 | 601.09 | 8.27 | 609.38 | 29.09 | -7.10 | 0.00 | 0.04 |
|  | … leaving after making a purchase [runtime] | 12,062.95 | 36.97 | 12,094.30 | 46.75 | 12,063.05 | 50.29 | -0.01 | 0.99 | 0.00 |
|  | … leaving before receiving normal help [runtime] | 1,133.80 | 67.36 | 1,343.30 | 63.92 | 1,706.55 | 99.01 | -21.39 | 0.00 | 0.14 |
|  | … leaving before receiving expert help [runtime] | 464.65 | 18.25 | 457.00 | 18.47 | 463.55 | 22.27 | 0.17 | 0.87 | 0.00 |
|  | … leaving whilst waiting to pay [runtime] | 7,048.70 | 105.59 | 7,520.30 | 116.47 | 7,633.60 | 124.13 | -16.05 | 0.00 | 0.02 |
|  | … leaving before finding anything [runtime] | 20,182.00 | 134.56 | 20,661.65 | 145.08 | 20,789.75 | 176.33 | -12.25 | 0.00 | 0.00 |
| WW | Overall customer count [per day] | 910.58 | 6.25 | 1,093.24 | 22.44 | 1,224.79 | 39.98 | -65.00 | 0.00 | 0.66 |
|  | … leaving after making a purchase | 29,696.75 | 44.48 | 30,097.55 | 31.34 | 30,256.45 | 45.29 | -39.43 | 0.00 | 0.03 |
|  | … leaving before receiving normal help | 5.40 | 3.30 | 29.55 | 7.24 | 84.60 | 17.71 | -19.66 | 0.00 | 0.81 |
|  | … leaving before receiving expert help | 83.60 | 9.04 | 130.75 | 12.48 | 170.40 | 14.53 | -22.68 | 0.00 | 0.67 |
|  | … leaving whilst waiting to pay | 6,291.95 | 165.73 | 13,148.10 | 200.16 | 18,050.20 | 164.96 | -224.88 | 0.00 | 0.68 |
|  | … leaving before finding anything | 27,662.80 | 166.79 | 33,120.80 | 198.49 | 37,173.90 | 196.79 | -164.89 | 0.00 | 0.30 |

Table 6: WOM descriptive statistics and t-tests between extreme conditions (AF = 0 versus AF = 1)

Given that we are looking at daily differences, it is important to examine what is going on graphically. Figure 5 presents time series data for a single model run which displays clear variation on a day-to-day basis (we have examined time series graphs of average daily performance, and the day-to-day variability which we wish to illustrate tends to average out over multiple model runs due to model stochasticity). As the adoption fraction increases, the number customer visits to WW generally also increases, which explains the unexpected increase in the count of customers leaving happy. This suggests that WW can cope well with the higher demand for service demanded by the additional customers who have been attracted through WOM. Results for A&TV suggest that it is not able to meet the increased customer demands which a higher adoption fraction places on it. The overall shape of the extreme case (adoption fraction 1) behaves in a pronounced cyclic manner with relatively large customer peaks and troughs; a sharp increase in one day's customers tends to be accompanied by a sharp decrease in the following day's customer numbers. Comparing the two departments, a greater increase in customer numbers can be seen between experimental conditions in WW than in A&TV as the customer pool grows, which is to be expected. In addition to the customer service limitations of A&TV, the starting customer pool size in WW is 23% greater than that of A&TV.

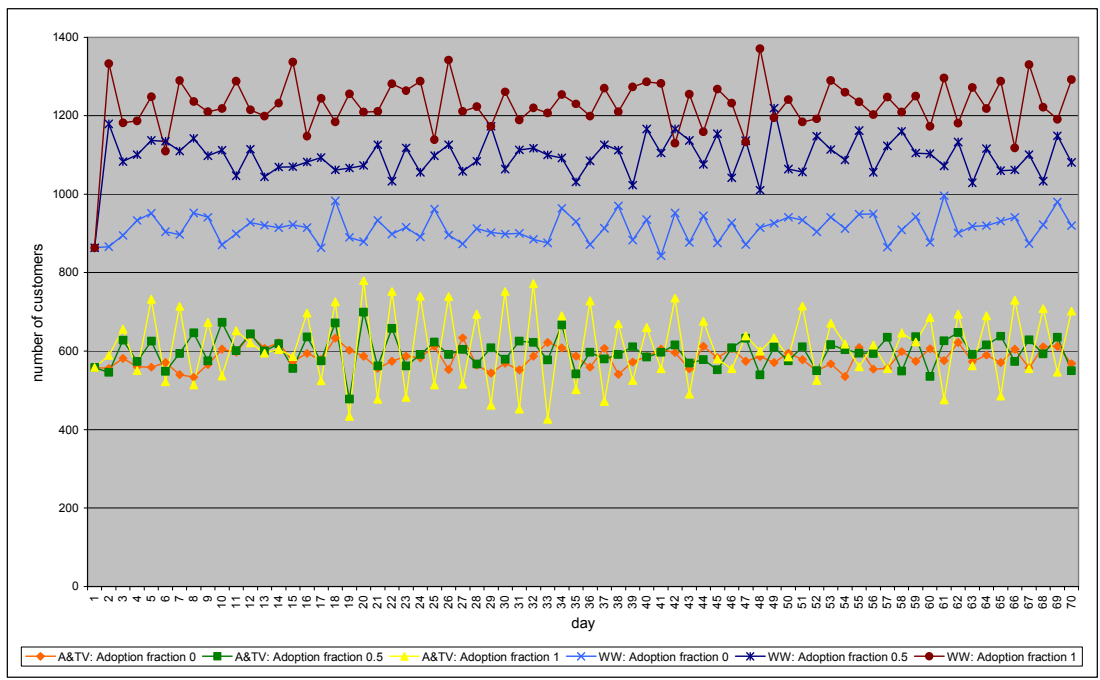

Figure 5: Overall number of customers per day (chart created from a single replication)

The experiment with the current WOM implementation has shown some interesting effects. In general our predictions have been borne out in our results, but more investigation is needed to fully understand the emergent patterns of performance. In the next version of our simulation model we will implement our second strategy to model the WOM phenomenon as described in section 4.2.2. This will provide us with more evidence to further confirm or disconfirm our current findings.

## 6. CONCLUSION

We have presented the conceptual design, implementation and operation of a simulation model that we are currently developing to help understand the impact of HR management practices on retail productivity. As far as we are aware this is the first time researchers have tried to use an agent-based approach to simulate management practices such as training and empowerment. Although our simulation model uses specific case studies as source of information, we believe that the general model could be adapted to other retail companies and areas of management practices that have a lot of human interaction.

In this paper we have focused in particular on the capabilities required to model customer evolution as a consequence of the implementation of management practices. We have discussed conceptual ideas about how to consider external and internal stimuli and have presented an implementation of WOM as one form of external stimuli. We are still testing and calibrating the new features we have implemented. The validation experiments so far have shown that we need to improve our simulation model in order to be able to model the real system in an appropriate way. In particular our current abstraction level with regards to how staff spend their time is much too high. If we want to use real staffing data we need to model how staff allocate their tasks between competing activities rather than focusing on one type of work. Overall the new features appear promising and we are convinced they will improve our insights into the operation of the departments within a department store. In particular the new performance measures we collect on a daily basis will be very useful in future for balancing services throughout the week.

In addition to continuing to validate our current simulation model we have also planned to experiment with more strategies of modelling customer evolution. We want to test a second WOM implementation where we expand or decrease our customer population as a consequence of the WOM spread rather than using the existing customers to model its influence. We will implement and test our conceptual ideas regarding customers' memory of their own shopping experiences (internal stimuli). Furthermore, we have planned to enhance the flexibility of staff members (i.e. empower them) to allow them to respond to customer demand. This will help to solve the staffing problem we have discussed above. In the long term we want to develop our simulation model to support testing the impact of team work related management practices. This looks like an interesting but challenging task because we first need to come up with a way to represent the effects of team work. Furthermore, we would like to enhance the capabilities of our agents, giving them skills in reasoning, negotiation, and co-operation.

Overall, we believe that researchers should become more involved in this multi-disciplinary kind of work to gain new insights into the behaviour of organisations. In our view, the main benefit from adopting this approach is the improved understanding of and debate about a problem domain. The very nature of the methods involved forces researchers to be explicit about the rules underlying behaviour and to think in new ways about them. As a result, we have brought work psychology and agent-based modelling closer together to form a new and exciting research area.

## REFERENCES

Birdi, K., Clegg, C.W., Patterson, M., Robinson, A., Stride, C.B., Wall, T.D., and Wood, S.J. (2006). "Contemporary Manufacturing Practices and Company Performance: A Longitudinal Study" Personnel Management, submitted.

Borshchev, A. and Filippov, A. (2004). "From System Dynamics and Discrete Event to Practical Agent Based Modeling: Reasons, Techniques, Tools" Proceedings of the 22nd International Conference of the System Dynamics Society, 25 - 29 July 2004, Oxford, England.

Celia, H. (2007). Retail Management Practices and Performance: On the Shop Floor. MA. University of the West of England.

Chick, S.E. (2006). "Six ways to improve a simulation analysis" Journal of Simulation, 1:21-28.

Delaney, J. T. and Huselid, M.A. (1996). "The Impact of Human Resource Management Practices on Performance in For-Profit and Non-Profit Organizations" Academy of Management Journal, 39:949-969.

Department of Trade and Industry (2003). "UK Productivity and Competitiveness Indicators" DTI Economics Paper No. 6.

Epstein, J.M. and Axtell, R. (1996). Growing Artificial Societies: Social Science from the Bottom Up. MIT Press: Cambridge, MA.

Fornell, C., Johnson, M.D., Anderson, E.W., Cha, J., and Bryant, B.E. (1996). "The American Customer Satisfaction Index: Nature, Purpose, and Findings" Journal of Marketing, 60(4):7-18.

Greasley, A. (2005). "Using DEA and Simulation in Guiding Operating Units to Improved Performance" Journal of the Operational Research Society, 56(6):727-731.

Hillier, F.S. and Lieberman, G.J. (2005). Introduction to Operations Research (8th Ed.). McGraw-Hill Higher Education: Boston, MA.

Hood, L. (1998). "Agent-based Modeling" Conference Proceedings: Greenhouse Beyond Kyoto, Issues, Op-portunities and Challenges, 31 March - 1 April 1998, Canberra, Australia.

Jeffrey, R. (2007). "Expert Voice: Icosystem's Eric Bonabeau on Agent-Based Modeling" Available via <http://www.cioinsight.com/article2/0,3959,1124316,00.asp>, [accessed 01/06/2007].

Keh, H.T., Chu, S., and Xu, J. (2006). "Efficiency, Effectiveness and Productivity of Marketing in Services" European Journal of Operational Research, 170(1):265-276.

Luck, M., McBurney, P., Shehory, O., and Willmott, S. (2005). Agent Technology: Computing as Interaction (A Roadmap for Agent Based Computing). AgentLink: Liverpool, England.

Marsden, P., Samson, A., and Upton, N. (2005). "The Economics of Buzz - Word of Mouth Drives Business Growth Finds LSE Study" Available via: <http://www.lse.ac.uk/collections/pressAndInformationOffice/newsAndEvents/archives/2005/Word_ofMouth.htm>, [accessed 29/11/2007].

Nicholson, M., Clarke, I., and Blakemore, M. (2002) "One Brand, Three Ways to Shop: Situational Variables and Multichannel Consumer Behaviour" International Review of Retail, Distribution and Consumer Research, 12:131-148.

Patel, S. and Schlijper, A. (2004). "Models of Consumer Behaviour". 49th European Study Group with Industry, 29 March - 2 April 2004, Oxford, England.

Pourdehnad, J., Maani, K., and Sedehi, H. (2002). "System Dynamics and Intelligent Agent-Based Simulation: Where is the Synergy?" Proceedings of the 20th International Conference of the System Dynamics Society, 28 July - 1 August 2002, Palermo, Italy.

Reynolds, J., Howard, E., Dragun, D., Rosewell, B., and Ormerod, P. (2005). "Assessing the Productivity of the UK Retail Sector" International Review of Retail, Distribution and Consumer Research, 15(3):237-280.

Robinson, S. (2004). Simulation: The Practice of Model Development and Use. John Wiley & Sons: Chichester

Schwaiger, A. and Stahmer, B. (2003). "SimMarket: Multi-Agent Based Customer Simulation and Decision Support for Category Management" Lecture Notes in Artificial Intelligence (LNAI) 2831. Springer: Berlin.

Shannon, R.E. (1975). Systems Simulation: The Art and Science. Prentice-Hall: Englewood Cliffs, NJ.

Siebers, P.O., Aickelin, U., Celia, H., and Clegg, C. (2007a). "A Multi-Agent Simulation of Retail Management Practices" Proceedings of the Summer Computer Simulation Conference (SCSC 2007), 15 - 18 July 2007, San Diego, USA.

Siebers, P.O., Aickelin, U., Celia, H., and Clegg, C. (2007b). "Using Intelligent Agents to Understand Management Practices and Retail Productivity" Proceedings of the Winter Simulation Conference (WSC 2007), 9 - 13 December 2007, Washington DC, USA.

Siebers, P.O., Aickelin, U., Celia, H., and Clegg, C. (2007c) "Understanding Retail Productivity by Simulating Management Practises" Proceedings of the EUROSIM Congress on Modelling and Simulation (EUROSIM 2007), 9 - 13 September, Ljubljana, Slovenia.

Simon, F. and Usunier, J.-C. (2007) "Cognitive, Demographic and Situational Determinants of Service Customer Preference for Personnel-in-Contact over Self-Service Technology". International Journal of Research in Marketing, 24(2):163-173.

Simon, H.A. (1996). The Sciences of the Artificial (3rd Ed.). MIT Press: Cambridge, MA.
Wall, T.D. and Wood, S.J. (2005). "Romance of Human Resource Management and Business Performance and the Case for Big Science" Human Relations, 58(5):429-462.
XJ Technologies (2007). "Official AnyLogic Website" Available via <http://www.xjtek.com/>. [accessed 01/06/2007].
Yi, Y. (1990) "A Critical Review of Consumer Satisfaction" Review of Marketing. American Marketing Association: Chicago, IL.

## AUTHOR BIOGRAPHIES

**PEER-OLAF SIEBERS** is a Research Fellow in the School of Computer Science and IT at the University of Nottingham. His main research interest is the application of computer simulation to study human oriented complex adaptive systems. Complementary fields of interest include distributed artificial intelligence, biologically inspired computing, game character behaviour modelling, and agent-based robotics. See his website for more details <www.cs.nott.ac.uk/~pos>.

**UWE AICKELIN** is a Reader and Advanced EPSRC Research Fellow in the School of Computer Science and IT at the University of Nottingham. His research interests are mathematical modelling, agent-based simulation, heuristic optimisation and artificial immune systems. See his webpage for more details <www.aickelin.com>.

**HELEN CELIA** is a Researcher at the Centre for Organisational Strategy, Learning and Change at Leeds University Business School. She is interested in developing ways of applying work psychology to better inform the modelling of complex systems using agents. See her website for more details <http://lubswww.leeds.ac.uk/lubs/index.php?id=103&backPID=95&tx_staffdetails_staff=275>

**CHRIS W. CLEGG** is a Professor of Organisational Psychology at Leeds University Business School and the Deputy Director of the Centre for Organisational Strategy, Learning and Change. His research interests include: new technology, systems design, information and control systems, socio-technical thinking and practice; organisational change, change management, technological change; the use and effectiveness of modern management practices, innovation, productivity; new ways of working, job design, and work organisation. An emerging research interest is modelling and simulation. See his webpage for more details <http://lubswww.leeds.ac.uk/lubs/index.php?id=103&backPID=95&tx_staffdetails_staff=274>